\documentclass[manuscript,screen]{acmart}

\AtBeginDocument{%
  \providecommand\BibTeX{{%
    \normalfont B\kern-0.5em{\scshape i\kern-0.25em b}\kern-0.8em\TeX}}}

\usepackage{bm}
\usepackage{tabularx}
\usepackage{textcomp}
\usepackage{xcolor}
\usepackage{multirow}
\usepackage{algorithm,algpseudocode}
\usepackage{caption}
\usepackage{wrapfig}

\theoremstyle{definition}
\newtheorem{definition}{Definition}

\setcopyright{acmcopyright}
\copyrightyear{2022} 
\acmYear{2022} 
\setcopyright{acmlicensed}\acmConference[EAAMO '22]{Equity and Access in Algorithms, Mechanisms, and Optimization}{October 6--9, 2022}{Arlington, VA, USA}
\acmBooktitle{Equity and Access in Algorithms, Mechanisms, and Optimization (EAAMO '22), October 6--9, 2022, Arlington, VA, USA}
\acmPrice{15.00}
\acmDOI{10.1145/3551624.3555287}
\acmISBN{978-1-4503-9477-2/22/10}

\begin{document}

\title{FairEGM: Fair Link Prediction and Recommendation via Emulated Graph Modification}

\author{Sean Current}
\email{current.33@osu.edu}
\orcid{0000-0002-3510-6919}
\affiliation{%
  \institution{Ohio State University}
  \streetaddress{281 W Lane Ave}
  \city{Columbus}
  \state{Ohio}
  \country{USA}
  \postcode{43210}
}

\author{Yuntian He}
\email{he.1773@osu.edu}
\affiliation{%
  \institution{Ohio State University}
  \streetaddress{281 W Lane Ave}
  \city{Columbus}
  \state{Ohio}
  \country{USA}
  \postcode{43210}
}

\author{Saket Gurukar}
\email{gurukar.1@osu.edu}
\affiliation{%
  \institution{Ohio State University}
  \streetaddress{281 W Lane Ave}
  \city{Columbus}
  \state{Ohio}
  \country{USA}
  \postcode{43210}
}
\author{Srinivasan Parthasarathy}
\email{srini@cse.ohio-state.edu}
\affiliation{%
  \institution{Ohio State University}
  \streetaddress{281 W Lane Ave}
  \city{Columbus}
  \state{Ohio}
  \country{USA}
  \postcode{43210}
}

% \renewcommand{\shortauthors}{Current et al.}

\begin{abstract}

% As machine learning becomes more widely adopted across domains, it is critical that researchers and ML engineers think about the inherent biases in the data that may be perpetuated by the model. Recently, many studies have shown that such biases are also imbibed in Graph Neural Network (GNN) models if the input graph is biased. In this work, we aim to mitigate the bias learned by GNNs through modifying the input graph. To that end, we propose \textbf{FairMod}, a \textbf{Fair} Graph \textbf{Mod}ification methodology with three formulations: the Global Fairness Optimization (GFO), Constrained Fairness Optimization (CFO), and Fair Edge Weighting (FEW) models. Our proposed models perform either microscopic or macroscopic edits to the input graph while training GNNs and learn node embeddings that are both accurate and fair under the context of link recommendations. We demonstrate the effectiveness of our approach on four real world datasets and show that we can improve the recommendation fairness by several factors at negligible cost to link prediction accuracy.

As machine learning becomes more widely adopted across domains, it is critical that researchers and ML engineers think about the inherent biases in the data that may be perpetuated by the model. Recently, many studies have shown that such biases are also imbibed in Graph Neural Network (GNN) models if the input graph is biased, potentially to the disadvantage of underserved and underrepresented communities. In this work, we aim to mitigate the bias learned by GNNs by jointly optimizing two different loss functions: one for the task of link prediction and one for the task of demographic parity. We further implement three different techniques inspired by graph modification approaches:  the Global Fairness Optimization (GFO), Constrained Fairness Optimization (CFO), and Fair Edge Weighting (FEW) models. These techniques mimic the effects of changing underlying graph structures within the GNN and offer a greater degree of interpretability over more integrated neural network methods. Our proposed models emulate microscopic or macroscopic edits to the input graph while training GNNs and learn node embeddings that are both accurate and fair under the context of link recommendations. We demonstrate the effectiveness of our approach on four real world datasets and show that we can improve the recommendation fairness by several factors at negligible cost to link prediction accuracy.

\end{abstract}


%   <concept>
%       <concept_id>10010147.10010257.10010293.10010294</concept_id>
%       <concept_desc>Computing methodologies~Neural networks</concept_desc>
%       <concept_significance>500</concept_significance>
%       </concept>
%  </ccs2012>

% \ccsdesc[500]{Computing methodologies~Neural networks}

\keywords{graph representation learning, demographic parity, group fairness, graph convolution neural networks, graph neural networks, graph fairness, link prediction, link recommendation}

\maketitle

\section{Introduction}
\label{sec:introduction}

% talk about gcns being used in the real world/making decisions for people
% highlight limitations/bias
% address our goals
% outline the methods we propose
% outline results
% outline contributions

The rapid development and widespread application of machine learning (ML) models in the day-to-day life of individuals highlights the pressing need to incorporate fairness in model formulation and development. An increasing number of such ML models operate on graph data. Of these ML models, graph neural networks (GNNs) \cite{kipf2016semi} have been shown to be effective on several machine learning tasks on graphs such as node classification \cite{perozzi2014deepwalk}, link prediction \cite{grover2016node2vec}, and graph visualization \cite{wang2016structural}. Due to their effectiveness, GNNs have been utilized in various applications such as recommendation engines \cite{pinsage2018}, drug discovery \cite{zitnik2017predicting}, and predicting social network relations \cite{kipf2016variational}.

\begin{figure}
    \centering
    \includegraphics[width=\textwidth]{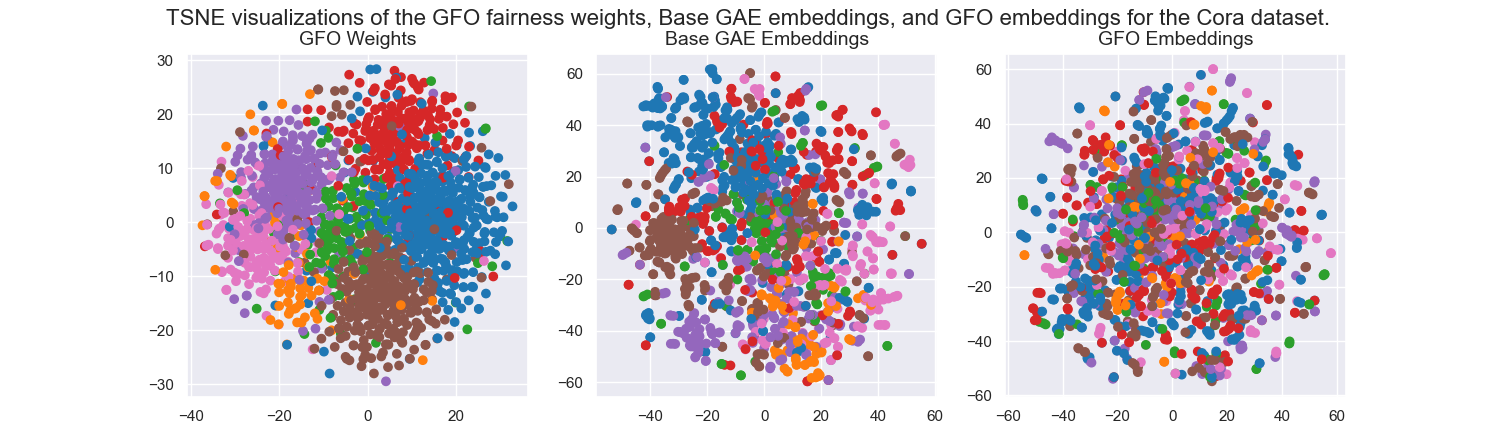}
    \caption{A TSNE\cite{vandermaaten08a} visualization of weights learned by the GFO method (Section \ref{sec:gfo}) as well as node embeddings produced by unfair and fair embedding methods on the Cora dataset. The color of a data-point indicates the value of the sensitive attribute for the corresponding node. Left: $\bm{W}_f$ weights learned by the GFO method. Middle: embeddings produced by a standard GAE autoencoder. Right: embeddings produced by the GFO embedding method. We observe that with the Base GAE embedding method, points with the same sensitive attribute are grouped more closely together, and thus links between points with the same sensitive attribute are more likely to be suggested by a link recommendation algorithm. In contrast, sensitive attributes in the GFO embeddings are more uniformly distributed, leading to more diversified recommendations with respect to the target sensitive attribute. This is achieved by learning weights which are added to nodes in the network to encourage fair representations; these weights inherently group according to the sensitive attribute features that they ``correct'' in order to make the node embeddings fair, as visualized in the left figure.}
    \label{fig:tsne}
\end{figure}

While GNNs demonstrate excellent performance on graph machine learning, these models can learn and propagate bias present in the input graph and amplify it. 
Given their prevalence, it is important to study and rectify biases these models learn from implicit or explicit biases in the input graph. For example, recent studies have shown that standard GNNs for downstream tasks like link prediction learn node representations that perform poorly on various fairness measures \cite{li2021dyadic,agarwal2021towards}. A common source of bias in this context is
the homophilic effect, which observes that nodes tend to associate with similar peers \cite{mcpherson2001birds} based on demographic characteristics such as race, ethnicity, sex, and religion. Homophily has recently been pointed out in the social sciences as a leading cause of the \textit{glass ceiling effect} \cite{avin2015homophily, stoica2018algorithmic}. More specifically, it can lead to unfair treatment of 
% minority groups
historically disadvantaged and underserved communities in multiple cases, including ranking \cite{karimi2018homophily}, social perception \cite{lee2019homophily}, and job promotion \cite{tesch1995promotion, clifton2019mathematical}. Apart from the structural properties of graphs, feature information in the graphs is also a source of bias. Even if a sensitive feature is excluded, bias can still be inherited from other features closely correlated with the sensitive one \cite{pedreshi2008discrimination}. Thus, it is paramount to explicitly model for and measure such sources of bias in downstream tasks like link prediction.

One can learn fair node representations and mitigate disparate sources of biases present in the input graph by incorporating fairness constraints in GNN training. Given the proliferation of graph-based recommendation models \citep{pinsage2018}, we incorporate fairness constraints through demographic parity \cite{gajane2017formalizing}. Informally, demographic parity seeks to ensure that each group with a particular sensitive attribute receives the positive outcome at the same rate as other groups with different values for the same sensitive attribute \cite{gajane2017formalizing}. For example, a professional networking website can utilize a GNN-based recommendation engine to recommend a job opening to a certain number of individuals. Demographic parity based recommendations could ensure that no individuals of a particular race, ethnicity, or gender are less likely to receive such a recommendation. The same can be said for social networking sites; if a recommendation algorithm is not fair under the definition of demographic parity, then the algorithm will suggest individuals with similar demographics to be linked, reinforcing preexisting ties and limiting diversity in a social setting. In contrast, an algorithm that is fair under the context of demographic parity would guarantee that recommendations are made with such diversity in mind. We also acknowledge that there exist contexts where biased recommendations that prioritize certain demographics can be beneficial (such as for the purpose of strengthening minority communities)\cite{fabbri2020effect}. In these instances, the application of demographic parity is not suitable; however, for the general level of link recommendations on a global scale, optimizing demographic parity can be a useful way to improve fairness and diversity in downstream applications.

We incorporate demographic parity in GNN-based recommendation models in an model-agnostic manner by proposing a novel \textit{Link Divergence} loss function that can be used in tandem with pre-existing link recommendation losses. We incorporate the optimization of Link Divergence through three separate methods inspired by graph modifications. Our techniques emulate the effect of adding new nodes and edges to the input graph without the computational overhead of actually doing so, offering both efficiency and interpretability. When used in tandem with the proposed Link Divergence loss, these methods facilitate the learning of fair graph node embeddings for downstream tasks of link prediction and recommendation. We present \textbf{E}mulated \textbf{G}raph \textbf{M}odifications for \textbf{Fair}ness (\textbf{FairEGM}), a collection of three methods which emulate the effects of a variety of graph modifications for the purpose of improving graph fairness. The Global Fairness Optimization (GFO) method introduces a fairness-oriented bias to every node in the dataset, which is optimized for fairness separately from the link prediction task. The Constrained Fairness Optimization (CFO) method introduces a rank-deficient weight matrix which is added as a bias to nodes in the graph, and is similarly optimized for fairness without regard to the link prediction task. Finally, the Fair Edge Weighting (FEW) method introduces edge weights to existing edges in the graph, and optimizes the edge weights to debias the input graph.

In our experiments, we concretely observe that applying models to biased data without regard to fairness can result in biased node representations that cluster based on the sensitive attribute under consideration, as shown in Figure \ref{fig:tsne}, middle. In contrast, our fair embedding models result in node representations that do not cluster by sensitive attribute (Figure \ref{fig:tsne}, right) while maintaining performance on downstream tasks by introducing corrective weights specific to nodes in the graph (Figure \ref{fig:tsne}, left). We show that each of or methods is capable of improving fairness across four real-world datasets and can significantly reduce the biases learned by GNNs. The reduced bias (or increased fairness) comes with a slight reduction in the link prediction performance. We also show that \textbf{FairEGM} is fairer than other graph learning baselines on the link prediction task.

% Standard GNNs learn unfair node representations under the context of fair link recommendation. Indeed, our results presented in Figure \ref{fig:training_metrics} in appendix APP show that as the link prediction utility metric (left) for a standard GNN autoencoder (Base) decreases, the unfairness metric increases (right). The utility measure and the unfairness measure here are captured through reconstruction \ref{eqn:recon} and link divergence loss \ref{eq:link_div}, respectively.
% We demonstrate each method on four real world networks, training graph neural networks for the link prediction task while simultaneously optimizing Link Divergence to balance the impact of different populations on each node and satisfy the fairness constraints.
% We perform experiments on four real-world datasets. We show that on all the datasets, our proposed models can significantly reduce the biases learned by GNNs. The reduced bias (or increased fairness) comes with a slight cost in the link prediction performance. We also show that \textbf{FairEGM} is fairer than other graph learning based baselines on the link prediction task.
% while also enhancing demographic parity on the link recommendation task.

\section{Related Work}
\label{sec:related}

\paragraph{Graph Representation Learning}

Various methods for graph representation learning for node embeddings have been proposed, such as Node2Vec \cite{grover2016node2vec}, DeepWalk \cite{perozzi2014deepwalk}, matrix factorization approaches \cite{GOYAL201878}, and graph neural network (GNN) approaches \cite{kipf2016semi}. Methods such as Node2Vec and DeepWalk construct embeddings using techniques inspired by word embeddings in natural language processing; these methods first generate random walks throughout the graph structure, forming sequences of nodes. These sequences are treated as ``sentences'', with the nodes as ``words'' in the sentence, allowing for the use of traditional word embedding algorithms to learn node embeddings. In contrast, matrix embedding approaches operate on the adjacency matrix of the graph $\bm{A}$, hoping to learn a set of embeddings $\bm{\Phi}$ such that $\bm{\Phi}\bm{\Phi^{\top}} = \bm{A}$.

However, when applied to attributed graphs, these methods are generally inferior to graph convolution networks. GCNs utilize the adjacency matrix $\bm{A}$ and a layer input $\bm{H}$ (usually the feature matrix $\bm{F}$ in the first layer of a GCN or the output of the previous layer $\bm{H}^{(\ell)}$) to learn node embeddings. Node embeddings $\bm{\Phi} = \bm{H}^{(\ell + 1)}$ are extracted from the final layer in the GCN, and are learned by directly optimizing a weight matrix $\bm{W}$ for the task such that
\begin{equation*}
    \bm{H}^{(\ell + 1)} = \sigma(\bm{D}^{-\frac{1}{2}}\bm{A}\bm{D}^{-\frac{1}{2}}\bm{H}^{(\ell)}\bm{W}),
\end{equation*}
where $\sigma$ is the sigmoid function (or some other nonlinear activation function, often ReLU) and $\bm{D} = \sum_{j}A_{ij}$ is a normalization vector for $A$. These methods have the added benefit of utilizing message-passing and semi-supervised learning in the embedding process, strengthening their performance compared to other methods on graph representation learning tasks. We will focus on GCNs as the primary embedding method for this paper.

\paragraph{Fairness in Machine Learning}

Recent work in machine learning has demonstrated the capability of ML models to learn implicit biases present in the data, such as systemic racism in criminal recidivism models \cite{angwin2016}, facial recognition technology \cite{buolamwini2018gender}, and gender biases in machine translation \cite{stanovsky2019evaluating}. As the usage of ML models across domains becomes increasingly common, it remains critical that researchers consider the fairness of their models.

Several formulations of fairness have been proposed in the ML community \cite{DBLP:journals/corr/abs-1908-09635}, \cite{gajane2017formalizing}. The first and most basic formulation is to ignore sensitive attributes with attribute unaware fairness: if a model cannot use a sensitive attribute in its prediction, the model is fair \cite{gajane2017formalizing}. However, this formulation is flawed due to correlations that may exist between sensitive attributes and other features in the dataset, such as zip-code when determining credit approval; due to systemic racism and historic segregation, some zip-codes are more strongly associated with specific races than others, which results in zip-code becoming a proxy for race.

This leads to the formulations of demographic parity and equalized odds \cite{gajane2017formalizing}, which propose specific constraints on the performance of a model. Demographic parity requires that members of different protected classes appear in the positive class at the same rate; the distribution of protected attributes of members in the positive class should match the population distribution. On the other hand, equalized odds is less focused on the model outcome, but rather on model performance; the true positive rates should be equal across protected attributes. This guarantees that a model achieves similar performance across protected attributes. In contrast to the prior definitions of fairness, the individual fairness \cite{gajane2017formalizing} definition does not depend on sensitive attributes, but rather the similarity of members. Individual fairness dictates that similar individuals should have similar outcomes in the model. 

%We focus on demographic parity; members of different protected classes should appear in the positive class at the same rate.

\paragraph{Fairness in Graph Embeddings}

The analysis of fairness in graph mining has received much attention in recent years. To address the bias in graph learning models, several models have been proposed \citep{kang2020inform,li2021dyadic,bose2019compositional,agarwal2021towards}. Bose et al. \cite{bose2019compositional} propose an adversarial method to ensure graph embeddings do not contain information that can be used to discern an individual's protected class. Rahman et al. \cite{ijcai2019-456} implement the FairWalk algorithm, which improves upon random walk algorithms by more fairly traversing the graph structure based upon the sensitive attributes of nodes.

In the realm of individual fairness, InFoRM \cite{kang2020inform} recognizes three approaches to implementing individual fairness constraints: debiasing the input graph, debiasing the mining model, and debiasing the mining result. To debias the input graph, Kang et al. construct an algorithm to optimize the adjacency matrix of graph data to improve individual fairness. Comparatively, the recent work of Li et al. \cite{li2021dyadic} proposes the FairAdj algorithm for group fairness metrics, which attempts to learn a fair adjacency matrix while preserving predictive accuracy for a dyadic link prediction task under structural constraints. Similarly, FairDrop \cite{spinelli2021biased} randomly drops edges in the graph while improving fairness based on dyadic attributes, also modifying the adjacency matrix.

Much of the previous work on debiasing the input graph focuses purely on modifying the existing adjacency matrix, in contrast to our work, which additionally considers debiasing the input graph by emulating the addition of new artificial nodes and edges. Not only is this approach unique, but it also allows for new information to be added to the network, contrasting approaches centered around optimizing the adjacency matrix, which are constricted to leveraging information already present in the graph. Additionally, there is opportunity to learn more about the features of the network through the analysis of added nodes: by understanding how the added information debiases the input graph, we can better understand biases present in the original graph.

\section{Problem Statement}
\label{sec:statement}

Our formulations consider an undirected graph $\bm{G} = (\bm{V}, \bm{E})$ with vertex set $\bm{V}$ containing $n$ nodes and edge set $\bm{E} \subseteq \bm{V} \times \bm{V}$. We notate the self-connected adjacency matrix $\bm{A}\in \mathbb{R}^{n \times n}$, the degree matrix $\bm{D}\in \mathbb{R}^{n \times n}$ and the normalized adjacency matrix $\hat{\bm{A}} = \bm{D}^{-\frac{1}{2}}\bm{A}\bm{D}^{-\frac{1}{2}}$. The node feature matrix is written as $\bm{F}\in \mathbb{R}^{n \times m}$, where $m$ is the dimensionality of the feature set, and the one-hot encoded sensitive attribute matrix as $\bm{S}\in \mathbb{N}^{n \times k}$ where $k$ is the number of possible values the sensitive attribute can take on (note that in the current formulation, only a single sensitive attribute with $k$ possible values is considered).
%\sg{Let $S^{(i)} \in \mathbb{R}^{n \times 1}$ denote the sensitive attribute vector for $i^{th}$ ( $0 \leq i \leq k$) sensitive attribute.}
The GNN weight matrix for layer $i$ is notated $\bm{W}^{(i)}\in \mathbb{R}^{l_{i-1} \times l_{i}}$, where $l_{i}$ is the hidden layer size and $l_{i-1}$ is the size of the previous layer ($m$ if $i=0$). The embedding matrix $\bm{\Phi}^{(i)}\in \mathbb{R}^{n \times l_{i}}$ is equal to the output of layer $i$ in the GNN. $\bm{\Phi}\in \mathbb{R}^{n \times d}$ is the output of the final layer of the GNN. %We refer to the weight matrix and embedding output for layer $i$ in the GNN as $\bm{W}^{(i)}$ and $\bm{\Phi}^{(i)}$, respectively.}
We further represent a generic GNN architecture as $\bm{\mathcal{G}}$.

\begin{definition}[Graph Representation Learning \citep{hamiltongrl}]
Given a graph $\bm{G} = (\bm{V}, \bm{E})$ with feature matrix $\bm{F}$ and a dimensionality $d \ll |V|$, graph representation learning aims to learn a function $h_{\bm{G}}: V \rightarrow \mathbb{R}^{d}$ such that 
%$\bm{\Phi} = h_{\bm{G}}(v)$ 
$\bm{\Phi} = h_{\bm{G}}(V)$
is a matrix of $d$-dimensional vector representations for nodes in the graph such that the similarity among nodes in the graph space is approximated by similarity between nodes in the embedding space.
\end{definition}

The learned node embeddings can be used as latent features in various downstream 
%\sg{machine learning} 
tasks. In this work, we focus on the fair link prediction task. We select demographic parity \cite{gajane2017formalizing} as our fairness criteria. Informally, demographic parity is satisfied if the output of the model is not dependent on a given sensitive attribute \cite{DBLP:journals/corr/abs-1908-09635}. Formally, we define the demographic parity fairness criteria on the link prediction problem as follows.

%consider link prediction and recommendation - the former aims to predict if two nodes are connected given their embeddings - while the latter aims to recommend links for each node in the graph. 

% \begin{definition}[Demographic Parity Fairness] 
% \sg{The probability of forming a link between two nodes $(u,v)$ should be independent of their sensitive attributes. Formally, $P(l(u,v)|S(u)=S(v)) = P(l(u,v)|S(u)!=S(v)) $}
% \end{definition}

\begin{definition}[Link Prediction with Demographic Parity Fairness]
\label{def:lp_dp}
Given a graph $\bm{G} = (\bm{V}, \bm{E})$, a node $v$, and an embedding model $M$, let $L_M(v) = (u_1, ..., u_n)$ be the set of nodes that have the highest likelihood to form a link with node $v$ where $L_M(v)$ is computed using the model $M$. The link prediction problem with demographic parity fairness for node $v$ has the following constraint: $D(P_{L_M(v)}, P_S)=0$ where $D$ is the distance metric between distributions, $P_{L_M(v)}$ is the distribution of sensitive attributes over the recommended nodes set $L_M(v)$, and $P_S$ is the distribution of sensitive attributes on the overall graph.
\end{definition}

The link prediction problem with demographic parity fairness states that the distribution of sensitive attributes over recommended nodes ($L_M(v)$) should not be distinguishable from the distribution of sensitive attributes on the overall graph ($P_S$). Note that the model can perform link recommendation by performing either K-nearest neighbors or by using a classifier.

\smallskip
\noindent
\textbf{Problem Statement:} We want to learn a graph neural network (GNN) model $M$ such that $M$ performs well on the link prediction task while satisfying the demographic parity fairness constraint (defined in  Definition \ref{def:lp_dp}).

% and the fairness constraint (in definition \ref{def:lp_dp}) is also satisfied.

\section{Methodology}
\label{sec:methodology}
To address the problem defined in Section \ref{sec:statement},
%the problem mentioned in the above problem statement, 
we train our proposed models with two objective functions.
%separate loss 
In the first objective function, we optimize the model for utility  (better performance on the link prediction task), while in the second objective function, we optimize the model for fairness (demographic parity based recommendations). Our objective functions are described as follows:

%so that the utility constraint (good performance on link prediction task) and the fairness constraint ( demographic parity based recommendations) are satisfied. Specifically, we train our proposed models with the following loss functions:

\paragraph{Utility objective}
% \paragraph{Reconstruction Loss:}
We use matrix reconstruction loss as our utility objective function due to its effectiveness in link prediction \cite{kipf2016variational} and recommendation systems \cite{ning2011slim, radhakrishnan2021simple}.
% Intuitively, nodes connected in the original graph should have similar node embeddings. 
% Using 
Specifically, following Kipf et al. \cite{kipf2016variational}, we use the sigmoid of the dot product between
%two nodes
node embeddings to reconstruct the matrix. This reconstruction task acts as a proxy for the link prediction task. We minimize the difference between the reconstructed and original adjacency matrix by optimizing the node embeddings.

% , we utilize a graph autoencoder (GAE, \cite{kipf2016variational}) reconstruction loss $L_{R}$ to measure the distance of the similarity matrix $\sigma(\bm{\Phi}\bm{\Phi}^{\top})$ of the node embeddings to the original adjacency matrix. 

Mathematically, our utility objective is described as follows:
%Specifically,
\begin{equation}
    L_{R}(\bm{\Phi}) = \bm{W}_{pos} \circ H(\bm{A}, \sigma(\bm{\Phi}\bm{\Phi}^{\top})),
    \label{eqn:recon}
\end{equation}
where $H$ is the binary cross entropy function and $\bm{W}_{pos}$ is an element-wise weighting term, defined as
\begin{equation}
    \bm{W}_{pos} = (\bm{A} \cdot \frac{\|\bm{V}\|^{2} - 2\|\bm{E}\|}{2\|\bm{E}\|} + (1 - \bm{A})).
\end{equation}
$\bm{W}_{pos}$ balances positive and negative edges in the graph, placing more weight on positive edges when the graph is sparse, and less weight on positive edges when it is highly connected.

\paragraph{Fairness objective}
%\paragraph{Link Divergence:}
In order to achieve demographic parity for the link prediction task, we require each group present in the sensitive attribute set to receive the positive outcome at the same rate. In other words, the distribution of sensitive attributes in the positive outcome should match the population distribution of sensitive attributes. In this case, we define a positive outcome as a positive prediction of a link between nodes: $\sigma(\bm{\Phi}_{v}\bm{\Phi}_{u}^{\top}) > 0.5$ indicates a positive outcome for node $u$ given node $v$. We first define a function $f(v)$ to evaluate the distribution of similarity scores for sensitive attributes in relation to the node $v \in \bm{V}$. Specifically, $f(v)$ computes the total similarity score for all nodes 
%$v$ 
$u \in \bm{V} \backslash \{v\}$
grouped by a specific sensitive attribute value $\bm{S}_{u}$ divided by the total number of nodes in that group. Based on this insight, we propose the link divergence loss function $L_{D}$ to measure the sum of KL-divergences between the population distribution $P_{S}$ of the sensitive attributes and $f(v)$ for each node:
\begin{equation}
    \label{eq:link_div}
    L_{D}(\bm{\Phi}) = \sum_{v \in \bm{V}} D_{KL} \left ( P_{S} \parallel f(v) \right ),
\end{equation}
where $D_{KL}$ is the KL-divergence function and
\begin{equation}
    f(v) = \frac{1}{\|\bm{V}\| - 1}\sum_{u \in \bm{V} \backslash \{v\}}\sigma(\bm{\Phi}_{v}\bm{\Phi}_{u}^{\top}) \cdot \bm{S}_{u},
\end{equation}
where $\bm{S}_{u}$ is a one-hot encoded vector representing the sensitive attribute of node $u$. Note that $f(v)$ is normalized before the KL-divergence is calculated. The $L_{D}$ loss function encourages weights to learn demographic parity: when $L_{D}$ is minimized, the two distributions $f(v)$ and $P_{s}$ are equal, directly enforcing demographic parity.

\paragraph{Graph Convolution:}
Each of our approaches uses a graph convolution operation with the symmetrically normalized adjacency matrix containing self-loops as the initial step. Given a node $v$ with neighbors $\mathcal{N}(v)$ and initial feature/embedding matrix $\bm{\Phi}^{(i)}$, the graph convolution operation is defined as the aggregative step
\begin{equation}
\label{eqn:node_wise}
    \bm{\Phi}_{v}^{(i+1)} = \sigma\left(\sum_{u \in
        \mathcal{N}(v) \cup \{ v \}} \frac{e_{u,v}}{\sqrt{\hat{d}_u
        \hat{d}_v}} \bm{\Phi}_v^{(i)} \cdot W^{(i)}\right)
\end{equation}
where $\bm{\Phi}_{v}^{(i+1)}$ is the embedding of node $v$ at layer $(i+1)$ and $\hat{d}_v$ = 1 + $sum_{u \in \mathcal{N}(v)} \cdot e_{u,v}$. Rewriting this operation in matrix form, we have
\begin{equation}
    \bm{\Phi}^{(i+1)} = \sigma(\hat{\bm{A}}\bm{\Phi}^{(i)}\bm{W}),
\end{equation}
where $\hat{\bm{A}}$ is the symmetrically normalized adjacency matrix containing self-loops. We use this representation when defining our formulations.

\paragraph{Optimization Process:} 
%\sg{TODO: Comment about fight between fairness vs utility loss. Rewrite based on algo. @Sean: Para should answer why we chose two seperate weights; why our optimization is designed this way.}
In our formulations, we want to constrain optimization of the reconstruction loss and link divergence to separate weight terms. To do this, each formulation constructs two optimization problems dependent on separate weight terms solved using a joint optimization. The reconstruction loss weights are only changed according to the gradient of the reconstruction loss and similarly the link divergence weights are changed only according to the gradient of link divergence.

Additionally, we introduce a hyper-parameter $\lambda_{f} \in (0, \infty)$ to weigh the impact of the fairness loss in the above optimization process. By scaling the gradients of the link divergence by a factor of $\lambda_{f}$, we retain a level of control over the impact of the fairness optimizations on the model. Naturally, due to inherent biases present in data, many models suffer performance losses when fairness is highly weighted \cite{kleinberg2018inherent}. As such, $\lambda_{f}$ allows one to balance the importance of fairness and utility in model application. Larger values of $\lambda_{f}$ will scale up the gradients of the link divergence loss, placing a larger weight on the fairness optimization during training, while lower values of $\lambda_{f}$ will let the optimization place more emphasis on the reconstruction loss. This optimization process is outlined in algorithm \ref{alg:optimize}.

% \sfc{Note that all methods introduce fairness into the first layer of a GNN prior to any following GNN layers. As such, all mathematical descriptions for the GFO, CFO, and FEW methods are depicted in the first layer of a GNN; adding additional layers onto the GNN does not change the formulation of the methodologies.}

\alglanguage{pseudocode}
\begin{algorithm}[t]
    \small
    \caption{Joint optimization procedure.}
    \label{alg:optimize}
    \begin{flushleft}
        % \vspace{-4mm}
        \textbf{Input}: node features $\bm{F}$, normalized adjacency matrix $\hat{\bm{A}}$, sensitive attribute matrix $\bm{S}$, GNN utility weights $\bm{W}$, GNN fairness weights $\bm{W}_{f}$, learning rate $\eta$, and fairness weight $\lambda_{f}$. 
    \end{flushleft}
    %\vspace{1mm}
    \begin{algorithmic}[1]
        \While {$\bm{W}$ or $\bm{W}_{f}$ has not converged}
        \State {Compute $L_{R}$ according to Eqn. \ref{eqn:recon}}
        \State {Set $g_{\bm{W}} \leftarrow \nabla_{\bm{W}}L_{R}$}
        \State {Set $\bm{W} \leftarrow \bm{W} - \eta \cdot \mathrm{Adam}(\bm{W}, g_{\bm{W}})$}\vspace{0.4em}
        \State {Compute $L_{D}$ according to Eqn. \ref{eq:link_div}}
        \State {Set $g_{\bm{W}_{f}} \leftarrow \lambda_{f} \cdot \nabla_{\bm{W}_{f}}L_{D}$}
        \State{Set $\bm{W}_{f} \leftarrow \bm{W}_{f} - \eta \cdot \mathrm{Adam}(\bm{W}, g_{\bm{W}})$}\vspace{0.4em}
        \EndWhile
    \end{algorithmic}
    \begin{flushleft}
        % \vspace{-4mm}
        \textbf{Output}: Fair node embeddings $\bm{\Phi} \leftarrow \mathrm{GNN}_{\bm{W},\bm{W}_{f}}(\bm{F}, \hat{\bm{A}})$ 
    \end{flushleft}
\end{algorithm}

\subsection{Global Fairness Optimization (GFO)}
\label{sec:gfo}
%\sg{In the GFO methodology, we assign each node ($v$) in a graph with a partner node ($v*$).  The partner node $v*$ attributes are  initialized with the  attributes of node $v$ and the edge weight between ($v, v*$) is initialized to 1.}
We first consider emulating a graph modification resulting from the introduction of a new partner node for each node in the graph, which is only connected to the node it is partnered with. On a conceptual level, the partner node $v^{*}$ is responsible for balancing out biases present in the original node $v$.

We introduce a new set of $n$ nodes $\bm{V}^{*}$ with accompanying $n$ edges $\bm{E}^{*} = \{(v_{i}, v^{*}_{i})\}_{i=1}^{n}$ each with edge weight $1$ and features $\bm{F}^{*}$ initialized from a Glorot normal distribution \cite{pmlr-v9-glorot10a}. We add $\bm{V}^{*}$ and $\bm{E}^{*}$ to the original graph $\bm{G}$ to construct a modified graph $\tilde{\bm{G}}=(\bm{V}\cup\bm{V}^{*}, \bm{E}\cup\bm{E}^{*})$ 
with feature matrix
    $\tilde{\bm{F}} = \left[ \begin{array}{l}
    \bm{F} \\
    \bm{F}^{*}
    \end{array} \right]$ 
and adjacency matrix
    $\tilde{\bm{A}}= \left[ \begin{array}{l l}
    \hat{\bm{A}} & \bm{A}^{*} \\
    \bm{A}^{*} & 0
    \end{array}\right]$,
where $\bm{A}^{*}$ is the $n \times n$ matrix with edge weights on the diagonal and 0 elsewhere.
%Note that for the GFO approach, $\bm{A}^{*}$ is initialized to the $n \times n$ identity matrix $\bm{I}^{n \times n}.$

Applying the graph convolution operation on the modified graph $\tilde{\bm{G}}$, we obtain node embeddings %following the first GNN layer
\begin{equation}
    \tilde{\bm{\Phi}}
    = \left[ \begin{array}{l}
    \bm{\Phi} \\
    \bm{\Phi}^{*}
    \end{array}\right]
    = \mathcal{G}(\sigma(\tilde{\bm{A}}\tilde{\bm{F}}\bm{W})),
    \label{eq:GFO_augmented}
\end{equation}
%\sg{TODO: the above assumes single GCN layer. Generalize it to L number of layers or mention that it is easily generalizable. Need to discuss the jump from eq. 5 to 6.}
where $\sigma$ is a nonlinear activation and $\mathcal{G}$ represents subsequent GNN layers. Since we do not need to learn resulting node embeddings for the introduced artificial nodes, we can simplify Equation \ref{eq:GFO_augmented} to only calculate embeddings for nodes in the original graph:
\begin{equation}
    \bm{\Phi} = \mathcal{G}(\sigma(\left[\hat{\bm{A}}~\bm{A}^{*}\right]\tilde{\bm{F}}\bm{W})) = \mathcal{G}(\sigma((\hat{\bm{A}}\bm{F} + \bm{A}^{*}\bm{F}^{*})\bm{W})).
    \label{eq:GFO_intermediate}
\end{equation}
We can further simplify the equation by distributing the edge weights of $\bm{A}^{*}$ to construct the fairness optimization weights $\bm{W}_{f} = \bm{A}^{*}\bm{F}^{*}$. Substituting $\bm{W}_{f}$ into Equation \ref{eq:GFO_intermediate}, we obtain the final GFO embedding formulation
\begin{equation}
    \bm{\Phi} = \mathcal{G}(\sigma((\hat{\bm{A}}\bm{F} + \bm{W}_{f})\bm{W})).
    \label{eq:GFO_final}
\end{equation}
Note that since there are no constraints on the values of $\bm{W}_{f}$, the GFO formulation equates to direct modification of node features following the convolution operation.

This results in the following optimization problems, dependent on $\mathcal{G}$, $\bm{W}$ (weights for the utility objective), and $\bm{W}_{f}$ (weights for the fairness objective):
\begin{equation}
    \min_{\mathcal{G},\bm{W}}(L_{R}(\bm{\Phi})) \text{ and } \min_{\bm{W}_{f}}(L_{D}(\bm{\Phi}))
\end{equation}

\subsection{Constrained Fairness Optimization (CFO)}
\label{sec:cfo}

Next, we consider a generalization of the GFO approach. Here, instead of introducing a partner node for every node in the graph, we introduce a finite set of $c$ new nodes connected to each node in the original graph. On a conceptual level, the $c$ new nodes form a basis by which the biases of nodes can be corrected through new connections to a diverse set of nodes. We use the notation CFO$_{c}$ to represent the CFO method using $c$ additional nodes. 

We introduce $\bm{V}^{*}$ with $c$ nodes, $\bm{E}^{*} = \{(\bm{v}_{i}, \bm{v}^{*}_{j})\}_{i=1}^{n}\empty{}_{j=1}^{c}$ with $n \cdot c$ edges, and features $\bm{F}^{*}$. The features and edge weights are initialized with Glorot normal initialization \cite{pmlr-v9-glorot10a}. We add $\bm{V}^{*}$ and $\bm{E}^{*}$ to the original graph $\bm{G}$ to construct a modified graph $\tilde{\bm{G}}=(\bm{V}\cup\bm{V}^{*}, \bm{E}\cup\bm{E}^{*})$
with feature matrix
    $\tilde{\bm{F}} = \left[ \begin{array}{l}
    \bm{F} \\
    \bm{F}^{*}
    \end{array} \right]$ 
and adjacency matrix
    $\tilde{\bm{A}}= \left[ \begin{array}{l l}
    \hat{\bm{A}} & \bm{A}^{*} \\
    \bm{A}^{*\bm{\top}} & 0
    \end{array}\right]$,
where $\bm{A}^{*}$ is the $n \times c$ matrix of edge weights. 

Following the same formulation of the graph convolution operation as the GFO method, we obtain the following output of the first GNN layer for CFO:
\begin{equation}
    \bm{\Phi} = \mathcal{G}(\sigma((\hat{\bm{A}}\bm{F} + \bm{A}^{*}\bm{F}^{*})\bm{W})).
    \label{eq:CFO_final}
\end{equation}

%\sg{The node embedding computation formulation of CFO is similar to that GFO till Eqn. \ref{eq:GFO_intermediate} due to their equivalent augmentation strategy to the adjacency matrix $\hat{A}$ and features matrix $\bm{F}$. In CFO, since $c << n$, the matrix $\bm{A}^{*}\bm{F}^{*} \in \mathbb{R}^{n \times m}$ in Eqn. \ref{eq:GFO_intermediate} is guaranteed to be rank deficient as shown below.}
There are now two separate formulations that can take place depending on the number of added nodes $c$. If $c$ is less than both the number of original nodes $n$ and the number of features $m$, the $n \times m$ matrix $\bm{A}^{*}\bm{F}^{*}$ is guaranteed to be rank deficient:
\begin{align*}
    \text{rank}(\bm{A}^{*}\bm{F}^{*}) &\leq \min(\text{rank}(\bm{A}^{*}), \text{rank}(\bm{F}^{*})) \\
                                      &\leq \min(\min(n, c), \min(c, m)) \\
                                      &= c < \min(n, m).
\end{align*}
% \begin{equation}
%     \text{rank}(\bm{A}^{*}\bm{F}^{*}) \leq \min(\text{rank}(\bm{A}^{*}), \text{rank}(\bm{F}^{*})) \leq \min(\min(n, c), \min(c, m)) = c < \min(n, m).
% \end{equation}
% Because $\bm{A}^{*}\bm{F}^{*}$ is rank deficient, 

Due to the rank deficiency of $\bm{A}^{*}\bm{F}^{*}$, we cannot generalize the product ($\bm{A}^{*}\bm{F}^{*}$) into a single weight matrix ($\bm{W}_f$) as we did in Equation \ref{eq:GFO_final}. As a result, one needs to maintain the inherent constraints of rank deficiency during the optimization process in the CFO formulation. % there are inherent constraints placed on the system that must be preserved in this formulation. 
Hence, the matrices $\bm{A}^{*}$ and $\bm{F}^{*}$ must be optimized separately to ensure $\bm{A}^{*}\bm{F}^{*}$ cannot achieve full rank. This results in the following optimization problems, dependent on $\mathcal{G}$, $\bm{W}$, $\bm{A}^{*}$, and $\bm{F}^{*}$:
\begin{equation}
    \min_{\mathcal{G},\bm{W}}(L_{R}(\bm{\Phi})) \text{ and } \min_{\bm{A}^{*},\bm{F}^{*}}(L_{D}(\bm{\Phi}))
\end{equation}

In contrast, we can consider the case where $c$ is greater than or equal to either the number of original nodes $n$ or the number of features $m$. When $c \geq n$ or $c \geq m$, the rank of the matrix product $\bm{A}^{*}\bm{F}^{*}$ is limited by $\min(n, m)$. 
%\sg{In such a case, we can generalize the product ($\bm{A}^{*}\bm{F}^{*}$) to single weight matrix $W_f$.}
Because this formulation is unconstrained by the value of $c$, we can simplify Equation \ref{eq:CFO_final} to match the GFO solution (Equation \ref{eq:GFO_final}) by introducing the same weight matrix, $\bm{W}_{f} = \bm{A}^{*}\bm{F}^{*}$.
%  \begin{equation*}
    % \bm{\Phi} = \sigma((\hat{\bm{A}}\bm{F} + \bm{W}_{f})\bm{W}).
% \end{equation*}
Thus, we observe that the GFO formulation is a special case of CFO.
% method.

\subsection{Fair Edge Weighting (FEW)}
In both GFO and CFO methods, we mitigate the bias present in the input graph by emulating the introduction of new nodes. In the FEW method, we mitigate bias learned by the GCN by editing edge weights in the existing graph.
Edge weights in a graph act as a weighting function when learning the node embedding for a node $v$ from its neighbors $u \in \mathcal{N}(v)$. The edge weight for edge $(u,v)$ determines the degree to which the features of node $u$ contribute to the embedding of node $v$. By scaling the edge weights in the adjacency matrix, we can place more emphasis on particular edges that could correct the bias for nodes in the graph. On a conceptual level, FEW balances edge weights in the existing graph to correct the bias present in the input data.

% On a conceptual level,  in FEW, we strengthen or weaken a tie $e_{(u, v)}$ and the modified tie strength is responsible to address the bias present in the graph. }
        
% Instead of introducing new nodes as the GFO and CFO methods accomplish, the FEW method focuses on introducing weights to existing edges in order to construct a fairer graph structure. 

We introduce an edge weight matrix, $\bm{W}_{f}$, to modify the existing normalized adjacency matrix $\hat{\bm{A}}$ prior to the graph convolution operation:
\begin{equation}
    \bm{\Phi} = \mathcal{G}(\sigma((\hat{\bm{A}} \circ \bm{W}_{f})\bm{F}\bm{W})).
    \label{eq:FEW_final}
\end{equation}
Because the normalized adjacency matrix $\hat{\bm{A}}$ is expected to have values of 0 for non-existent edges, the element-wise multiplication operation will only introduce weights on existing edges. This results in the following optimization problems, dependent on $\mathcal{G}$, $\bm{W}$, and $\bm{W}_{f}$:
\begin{equation}
    \min_{\mathcal{G},\bm{W}}(L_{R}(\bm{\Phi})) \text{ and } \min_{\bm{W}_{f}}(L_{D}(\bm{\Phi}))
\end{equation}

Note that in this formulation of FEW, there are no constraints on $\bm{W}_{f}$, so introduced weights have a range of $(-\infty, \infty)$ and are not necessarily symmetric, allowing FEW to construct a directed graph.

\section{Experiments}
\label{sec:experiment}

\subsection{Dataset}
We conduct our experiments on a set of four real-world datasets (see Table \ref{tab:datasets} for details, including sensitive attribute information).  Edges in the Citeseer, Cora, and Pubmed datasets (\url{https://linqs.soe.ucsc.edu/data}) represent paper citations and edges in the Facebook-1684 ego-network dataset (\url{https://snap.stanford.edu/data/}) represent Facebook friendships. Citeseer, Cora, and Pubmed have bag-of-word feature vectors for each node while Facebook-1684 has anonymized features for each node representative of various attributes of a person's Facebook profile.

\begin{table*}[!h]
    \centering
    \begin{tabular}{|c|c c c c c|}
         \hline
         Dataset & Nodes & Edges & Features & Sensitive Attribute & Clustering Coefficient\\
         \hline
         Citeseer & 3,327 & 4,732 & 3,703 & Topic (6) & 0.2407 \\
         Cora & 2,708 & 5,278 & 1,433 & Topic (7) & 0.1426 \\
         Facebook & 786 & 14,024 & 317 & Gender (2) & 0.4757 \\
         Pubmed & 19,717 & 44,327 & 500 & Topic (3) & 0.0602 \\
         \hline
    \end{tabular}
    \captionsetup{justification=centering}
    \caption{Dataset statistics for the four experimental datasets. The number (n) next to the \\sensitive attribute label indicates how many values the sensitive attribute may take on.}
    \label{tab:datasets}
\end{table*}

\subsection{Experimental Setup}
For each dataset, we train a basic graph convolution network (GCN), a GCN with GFO optimization, two GCNs with $CFO_{10}$ ($c=10$) and $CFO_{100}$ ($c=100$) optimization, and a GCN with FEW optimization. All models use a two-layer GAE autoencoder\citep{kipf2016variational} as the GCN model. Following prior work \citep{kipf2016variational}, each GAE has a 32-dim hidden layer and a 16-dim embedding layer. We compare our methods to FairWalk \citep{ijcai2019-456} and FairAdj \citep{li2021dyadic} using similar parameters. We offer an additional experiment in Appendix \ref{app:aug}, which compares our models to the base GAE model using an augmented loss function $L(\bm{\Phi}) = L_{R}(\bm{\Phi}) + \lambda L_{D}(\bm{\Phi}).$

Link predictions models are trained using 20 randomized train-test splits. Following prior works~\cite{gurukar2019network} for link prediction tasks, we split the dataset into training and test data using the following procedure. We first randomly sample 20\% of edges and add them to the test set. From the subgraph consisting of the remaining edges, we extract its largest connected component as the training data. Finally, we remove the nodes that are not in the training data but are present in the test set from the test set.

Training hyperparameters are chosen separately for each model based on reconstruction loss optimization. All values of the fairness weight $\lambda$ are set to 1. Losses are optimized with an Adam optimizer \cite{kingma2017adam} with $\beta_{1}$ and $\beta_{2}$ kept at the default values of $0.9$ and $0.999$, respectively, while the learning rate $\alpha$ and number of training epochs are tuned according to the datasets. Hyperparameters for each dataset are listed in the Table \ref{tab:datasetparams}. Any hyperparameters not listed are kept at their default values from the source code.

Experiments for the Pubmed dataset are run on a high-performance computer with Intel Xeon E5-2680 v4 CPUs (128GB memory) and NVIDIA Tesla P100 GPUs (16GB memory).
All other datasets are run on a laptop with Intel Core i7-9750H CPUs (16GB memory) and an NVIDIA GeForce GTX 1650 GPU (4GB memory).

\begin{table*}[!b]
    \centering
    \begin{tabular}{|c|c c|c c|c c|}
         \hline
         \multirow{2}{*}{Dataset} & \multicolumn{2}{c|}{GCN Models} & \multicolumn{2}{c|}{FairAdj}  &\multicolumn{2}{c|}{FairWalk} \\
         & Learning Rate & Epochs & Learning Rate & T2 & Learning Rate & Epochs\\
         \hline
         Citeseer & 0.0001 & 300 & 0.005 & 10 & 0.01 & 1 \\
         Cora     & 0.0001 & 300 & 0.001 & 10 & 0.1 & 1 \\
         Facebook & 0.0001 & 300 & 0.01  & 10 & 0.1 & 1 \\
         Pubmed   & 0.001 & 200 & 0.005 & 10 & 0.1 & 1 \\
         \hline
    \end{tabular}
    \caption{Model hyperparameters for the four experimental datasets.}
    \label{tab:datasetparams}
\end{table*}

\subsection{Metrics}
\noindent {\bf Loss:} We evaluate all methods based on reconstruction loss $L_R$ and link divergence $L_D$, as discussed previously.

\noindent {\bf Quality:} To evaluate  embedding quality on the link prediction task, we use the AUROC and F1-Score metrics following the procedures described in \cite{gurukar2019network}. We train a logistic regression model as a classifier for positive/negative edges using an equal number of positive and negative edges randomly selected from the training set. The AUROC and F1-Score are then recorded using the logistic regression predictions for positive edges and an equal number of randomly selected negative edges from the test set.

\noindent {\bf Fairness:}
To evaluate performance on the demographic parity fairness task for link recommendation, we compute DP@$k$ as follows: for each node $u$ in graph $\bm{G}$, we utilize the learned $d$-dimension embedding vector to find the $k$-nearest nodes (denoted as $\mathrm{kNN}(u)$) in the embedding space using the sigmoid of the dot-product similarity score. From the $k$-nearest nodes $\mathrm{kNN}(u)$, we calculate the distribution $\pi(\mathrm{kNN}(u))$ of the observed sensitive attributes and compare to the global distribution $P_{S}$ of sensitive attributes of the dataset. The metric is defined for a node $u$ as follows:
\begin{equation}
    \mathrm{DP@}k(u) = D_{KL} \left ( P_{S} \parallel \pi(\mathrm{kNN}(u)) \right )
    \label{eqn:dpku}
\end{equation}
where $\pi(\mathrm{kNN}(u))$ represents the normalized distribution of sensitive attribute values in the nearest neighbors of $u$ and $P_{S}$ is the distribution of the sensitive attributes in the overall dataset. The final overall metric DP@$k$ is the average DP@$k(u)$ for all nodes $u$ in the dataset:
\begin{equation}
    \mathrm{DP@}k = \frac{1}{|V|}\sum_{u}\mathrm{DP@}k(u).
    \label{eqn:dpk}
\end{equation}
Ideally, this value should be as close to zero as possible. When $S$ is a binary sensitive attribute, the DP@$k$ metric is additionally a suitable metric for dyadic fairness, and is similar to the $\Delta$DP metric proposed by \cite{li2021dyadic}.

% \sfc{We further evaluate models with a metric oriented toward dyadic fairness. To calculate DyF@$k\%$, we take the top $k$ percent of link predictions and compare the distribution of sensitive attribute pairs for each edge in the top $k$ percent to the expected distribution $P_{S,S}$ of sensitive attribute pairs if all edges between nodes were equally likely. Mathematically, we define DyF@$k\%$ as follows:}
% \begin{equation}
%     \mathrm{DyF@}k\%(\sigma(\bm{\Phi}\bm{\Phi}^{\top})) =
%     D_{KL} \left ( P_{S,S} \parallel \pi\left((s_{i}, s_{j})~:~ i < j ~\&~ (i, j) \in \sigma(\bm{\Phi}\bm{\Phi}^{\top})_{(100 - k)\%}\right) \right ).
% \end{equation}
% \sfc{Where $P_{S,S}$ is the expected distribution of sensitive attribute pairs, $s_{i}$ is the sensitive attribute of node $i$, $s_{j}$ is the sensitive attribute of node $j$, $(i, j)$ is the edge between node $i$ and node $j$, and $\sigma(\bm{\Phi}\bm{\Phi}^{\top})_{(100 - k)\%})$ is the top $k$ percent of edges predicted by $\sigma(\bm{\Phi}\bm{\Phi}^{\top})$. Ideally, this metric should be as close to zero as possible.}

\begin{table}[!t]
    \centering
    \footnotesize
    \begin{tabular}{l l l l l l}
        \toprule
        Dataset & Model & $L_{R}$ $\downarrow$ & $L_{D}$ $\downarrow$ & AUROC $\uparrow$ & F1 $\uparrow$ \\
        \toprule
        \multirow{7}{*}{Citeseer} & Base       & 1.31       $\pm$ 0.0399     & 0.000743   $\pm$ 0.00247    & \textbf{0.74}       $\pm$ 0.0295     & 0.616      $\pm$ 0.0259     \\
        \cmidrule{2-6}
& FairWalk   & 6.05       $\pm$ 0.308      & 0.000134   $\pm$ 3.47e-05   & 0.697      $\pm$ 0.0302     & \textbf{0.628}      $\pm$ 0.0366     \\
& FairAdj    & \textbf{0.832}      $\pm$ 0.00447    & 0.00227    $\pm$ 0.00021    & 0.678      $\pm$ 0.0262     & 0.596      $\pm$ 0.0264     \\
        \cmidrule{2-6}
& GFO        & 1.34       $\pm$ 0.0142     & \textbf{3.12e-08}   $\pm$ 4.74e-08   & 0.714      $\pm$ 0.0263     & 0.599      $\pm$ 0.0263     \\
& CFO$_{10}$ & 1.34       $\pm$ 0.0125     & 9.88e-07   $\pm$ 9.14e-07   & 0.714      $\pm$ 0.0339     & 0.599      $\pm$ 0.0354     \\
& CFO$_{100}$ & 1.34       $\pm$ 0.014      & 6.71e-08   $\pm$ 6.66e-08   & 0.706      $\pm$ 0.0544     & 0.595      $\pm$ 0.0339    \\
& FEW        & 1.33       $\pm$ 0.0242     & 0.000124   $\pm$ 0.000161   & 0.718      $\pm$ 0.0579     & 0.608      $\pm$ 0.0345     \\

        \toprule
        \multirow{7}{*}{Cora} & Base       & 1.29       $\pm$ 0.0188     & 0.000117   $\pm$ 0.000159   & 0.73       $\pm$ 0.0205     & 0.608      $\pm$ 0.0199     \\
        \cmidrule{2-6}
& FairWalk   & 6.89       $\pm$ 0.135      & 0.000453   $\pm$ 6.55e-05   & 0.624      $\pm$ 0.0158     & 0.573      $\pm$ 0.0141     \\
& FairAdj    & \textbf{0.937}      $\pm$ 0.0109     & 0.00547    $\pm$ 0.000376   & 0.573      $\pm$ 0.0309     & 0.554      $\pm$ 0.0215     \\
        \cmidrule{2-6}
& GFO        & 1.31       $\pm$ 0.00939    & \textbf{1.04e-07}   $\pm$ 6.5e-08    & \textbf{0.731}      $\pm$ 0.0182     & \textbf{0.613}      $\pm$ 0.0183     \\
& CFO$_{10}$ & 1.3        $\pm$ 0.0155     & 7.71e-06   $\pm$ 5.44e-06   & \textbf{0.731}      $\pm$ 0.0204     & \textbf{0.613}      $\pm$ 0.019      \\
& CFO$_{100}$ & 1.3        $\pm$ 0.0116     & 4.88e-07   $\pm$ 7.19e-07   & 0.729      $\pm$ 0.0207     & 0.609      $\pm$ 0.022     \\
& FEW        & 1.29       $\pm$ 0.0188     & 0.000101   $\pm$ 0.000112   & 0.726      $\pm$ 0.0226     & 0.607      $\pm$ 0.0242     \\

        \toprule
        \multirow{7}{*}{Facebook} & Base       & 1.25       $\pm$ 0.0295     & 1.43e-05   $\pm$ 1.65e-05   & 0.786      $\pm$ 0.0117     & 0.72       $\pm$ 0.0116     \\
        \cmidrule{2-6}
& FairWalk   & 2.26       $\pm$ 0.0333     & 0.000182   $\pm$ 3.07e-05   & 0.723      $\pm$ 0.00861    & 0.693      $\pm$ 0.00743    \\
& FairAdj    & \textbf{0.837}      $\pm$ 0.00208    & 0.000686   $\pm$ 5.58e-05   & 0.759      $\pm$ 0.00851    & 0.711      $\pm$ 0.00733    \\
        \cmidrule{2-6}
& GFO        & 1.27       $\pm$ 0.0189     & 6.89e-08   $\pm$ 5.99e-08   & \textbf{0.788}      $\pm$ 0.0094     & 0.72       $\pm$ 0.00808    \\
& CFO$_{10}$ & 1.26       $\pm$ 0.0185     & 4.45e-07   $\pm$ 3.96e-07   & 0.786      $\pm$ 0.00998    & 0.72       $\pm$ 0.00948    \\
& CFO$_{100}$ & 1.26       $\pm$ 0.0177     & \textbf{3.01e-08}   $\pm$ 1.66e-08   & 0.787      $\pm$ 0.00966    & \textbf{0.721}      $\pm$ 0.00699   \\
& FEW        & 1.24       $\pm$ 0.0409     & 1.65e-05   $\pm$ 2.34e-05   & 0.787      $\pm$ 0.00998    & \textbf{0.721}      $\pm$ 0.00918    \\

        \toprule
        \multirow{5}{*}{Pubmed} & Base       & 1.31       $\pm$ 0.00417    & 2.95e-06   $\pm$ 2.72e-06   & \textbf{0.847}      $\pm$ 0.00431    & \textbf{0.724}      $\pm$ 0.00499    \\
        \cmidrule{2-6}
& FairWalk   & 8.58       $\pm$ 0.0568     & 0.000302   $\pm$ 1.92e-05   & 0.724      $\pm$ 0.00612    & 0.622      $\pm$ 0.00585    \\
& FairAdj    & \textbf{0.875}      $\pm$ 0.00224    & 0.0045     $\pm$ 0.000141   & 0.63       $\pm$ 0.00934    & 0.562      $\pm$ 0.0104     \\
        \cmidrule{2-6}
& GFO        & 1.31       $\pm$ 0.0179     & 2.22e-09   $\pm$ 3.17e-09   & 0.829      $\pm$ 0.0717     & 0.716      $\pm$ 0.0314     \\
& CFO$_{10}$ & 1.31       $\pm$ 0.0181     & 4.56e-09   $\pm$ 3.44e-09   & 0.83       $\pm$ 0.0758     & 0.718      $\pm$ 0.0323     \\
& CFO$_{100}$ & 1.32       $\pm$ 0.0242     & 1.84e-09   $\pm$ \textbf{1.5e-09}    & 0.812      $\pm$ 0.104      & 0.711      $\pm$ 0.0444    \\
& FEW        & 1.31       $\pm$ 0.0176     & 5.41e-08   $\pm$ 7.84e-08   & 0.834      $\pm$ 0.053      & 0.716      $\pm$ 0.0318     \\

\bottomrule
    \end{tabular}
    \captionsetup{justification=centering}
    \caption{Utility in Link Prediction: losses and link prediction metrics for all datasets.\\ The highest performing model for each dataset and metric is bolded.}
    \label{tab:results}
\end{table}

\subsection{Results}
\label{sec:results}

Results for link prediction are reported in Table \ref{tab:results}. Across the board, our methods consistently perform near or above state-of-the-art levels for link prediction while additionally optimizing for link divergence. While FairAdj reports a lower reconstruction loss for all datasets, the AUROC and F1-scores do not behave similarly. For the Citeseer dataset, the AUROC and F1-scores are slightly better for FairAdj compared to the GFO, CFO$_{10}$, CFO$_{100}$ and FEW models, while the reverse is true for Cora, Facebook, and Pubmed, which report the fair autoencoder models scoring significantly higher on AUROC, particularly for the Cora and Pubmed datasets. We further note that the GFO, CFO$_{10}$, CFO$_{100}$, and FEW models do not consistently perform significantly better or worse than the base model they are built on, indicating that the fairness optimizations made by the models do not seem to significantly impact embedding performance.

Results for demographic parity in link recommendation with the DP@$k$ metrics are presented in Table \ref{tab:results2}. We observe that for the Citeseer dataset, the GFO method offers the greatest improvement in the DP@$k$ metrics compared to the base GCN model while maintaining similar AUROC and F1 scores. All of the fair autoencoders improve upon the base GCN method. This is similarly true for the Cora, Facebook, and Pubmed datasets, where the GFO method consistently ranks among the top models for the DP@$k$ metrics, though is arguably out-shined by the CFO$_{100}$ method for the Cora dataset. We additionally note that as $k$ increases, the DP@$k$ metrics decrease for all methods, indicating that as more of the nearest nodes are considered, the distribution of sensitive attributes becomes fairer and more representative of the population.

\begin{wraptable}{r}{9.cm}
    \centering
    \footnotesize
    \begin{tabular}{l l l l l}% l l}
        \toprule
        Dataset & Model & DP@10 $\downarrow$ & DP@20 $\downarrow$ & DP@40 $\downarrow$ \\
        \toprule
        \multirow{7}{*}{Citeseer} & Base       & 4.88       $\pm$ 1.79       & 3.01       $\pm$ 1.69       & 1.85       $\pm$ 1.4        \\
        \cmidrule{2-5}
& FairWalk   & 6.96       $\pm$ 0.79       & 5.31       $\pm$ 0.742      & 2.9        $\pm$ 0.557      \\
& FairAdj    & 6.54       $\pm$ 0.747      & 4.55       $\pm$ 0.762      & 2.16       $\pm$ 0.445      \\
        \cmidrule{2-5}
& GFO        & 2.3        $\pm$ 1.23       & \textbf{0.706}      $\pm$ 0.506      & \textbf{0.172}      $\pm$ 0.201      \\
& CFO$_{10}$ & \textbf{1.94}       $\pm$ 0.791      & 0.879      $\pm$ 0.527      & 0.234      $\pm$ 0.285      \\
& CFO$_{100}$ & 2.34       $\pm$ 1.17       & 0.71       $\pm$ 0.692      & 0.271      $\pm$ 0.516      \\
& FEW        & 4.17       $\pm$ 1.35       & 2.2        $\pm$ 1.28       & 1.17       $\pm$ 0.937      \\

        \toprule
        \multirow{7}{*}{Cora} & Base       & 3.79       $\pm$ 1.06       & 1.89       $\pm$ 0.934      & 0.914      $\pm$ 0.652      \\
        \cmidrule{2-5}
& FairWalk   & 7.35       $\pm$ 0.137      & 5.4        $\pm$ 0.208      & 3.37       $\pm$ 0.312      \\
& FairAdj    & 7.09       $\pm$ 0.189      & 4.73       $\pm$ 0.175      & 2.51       $\pm$ 0.167      \\
        \cmidrule{2-5}
& GFO        & 3.47       $\pm$ 0.826      & 1.05       $\pm$ 0.348      & \textbf{0.132}      $\pm$ 0.0486     \\
& CFO$_{10}$ & 3.38       $\pm$ 0.84       & 1.23       $\pm$ 0.536      & 0.239      $\pm$ 0.123      \\
& CFO$_{100}$ & \textbf{3.32}       $\pm$ 0.808      & \textbf{0.96}       $\pm$ 0.424      & 0.147      $\pm$ 0.0713     \\
& FEW        & 3.81       $\pm$ 0.874      & 1.87       $\pm$ 0.618      & 0.719      $\pm$ 0.42       \\

        \toprule
        \multirow{7}{*}{Facebook} & Base       & 0.136      $\pm$ 0.161      & 0.0418     $\pm$ 0.0204     & 0.0226     $\pm$ 0.0102     \\
        \cmidrule{2-5}
& FairWalk   & 0.239      $\pm$ 0.0764     & 0.0451     $\pm$ 0.00856    & 0.0226     $\pm$ 0.00254    \\
& FairAdj    & 0.411      $\pm$ 0.114      & 0.0802     $\pm$ 0.0264     & 0.0329     $\pm$ 0.00335    \\
        \cmidrule{2-5}
& GFO        & \textbf{0.0661}     $\pm$ 0.0484     & \textbf{0.0191}     $\pm$ 0.0114     & 0.0076     $\pm$ 0.00446    \\
& CFO$_{10}$ & 0.17       $\pm$ 0.329      & 0.0226     $\pm$ 0.00984    & \textbf{0.0072}     $\pm$ 0.00366    \\
& CFO$_{100}$ & 0.101      $\pm$ 0.0643     & 0.0246     $\pm$ 0.0133     & 0.00783    $\pm$ 0.00441    \\
& FEW        & 0.217      $\pm$ 0.391      & 0.0413     $\pm$ 0.0252     & 0.0171     $\pm$ 0.00914    \\

        \toprule
        \multirow{5}{*}{Pubmed} & Base       & 2.43       $\pm$ 1.36       & 0.64       $\pm$ 1.09       & 0.185      $\pm$ 0.0859     \\
        \cmidrule{2-5}
& FairWalk   & \textbf{1.12}       $\pm$ 0.0701     & 0.368      $\pm$ 0.0405     & 0.176      $\pm$ 0.0184     \\
& FairAdj    & 5.72       $\pm$ 0.108      & 4.46       $\pm$ 0.136      & 3.3        $\pm$ 0.148      \\
        \cmidrule{2-5}
& GFO        & 1.52       $\pm$ 1.56       & 0.327      $\pm$ 0.72       & 0.196      $\pm$ 0.0851     \\
& CFO$_{10}$ & 2.61       $\pm$ 1.27       & \textbf{0.174}      $\pm$ 0.0407     & 0.203      $\pm$ 0.0694     \\
& CFO$_{100}$ & 2.06       $\pm$ 1.51       & 0.182      $\pm$ 0.0322     & \textbf{0.168}      $\pm$ 0.0906     \\
& FEW        & 2.31       $\pm$ 1.38       & 0.328      $\pm$ 0.697      & 0.196      $\pm$ 0.0686     \\
        \bottomrule
    \end{tabular}
    \captionsetup{justification=centering}
    \caption{Fairness in Link Recommendations: DP@$k$ metrics for all datasets. The highest performing model for each dataset and metric is bolded.}
    \label{tab:results2}
\end{wraptable}

Additionally, our models perform better than Fairwalk and FairAdj for the Citeseer, Cora, and Facebook datasets, and better than FairAdj for the Pubmed dataset. Fairwalk achieves generally stronger DP@$k$ scores on the Pubmed dataset; however, this performance is offset by weaker AUROC and F1 scores, representative of the inherent trade-off between fairness and utility.

\begin{figure*}[!t]
    \centering
    \includegraphics[width=\textwidth]{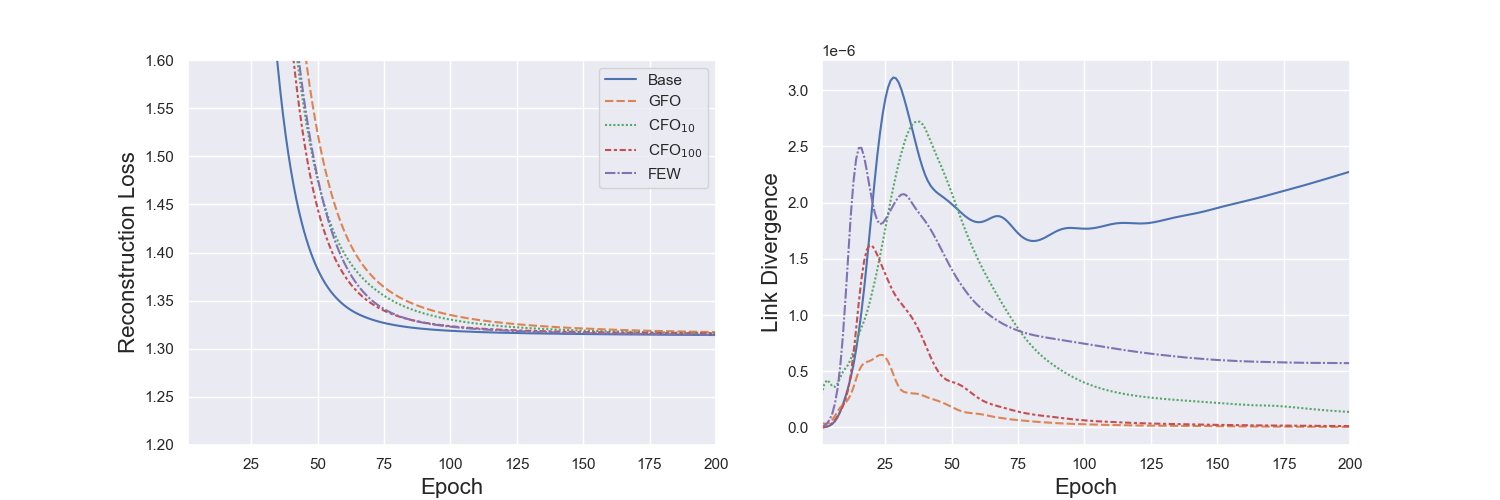}
    \caption{Left: Reconstruction loss during training for the Pubmed dataset. Right: Link divergence loss during training for the Pubmed dataset.}
    \label{fig:training_metrics}
\end{figure*}

The FEW method does not appear as capable of improving fairness as the GFO and CFO methods. This is further observed in Figure \ref{fig:training_metrics}, which documents the reconstruction and link divergence losses during training for the various methods on the Pubmed dataset. As the FEW model continues training, Link Divergence asymptotically converges to a higher value than the GFO and CFO methods. In contrast, the GFO and CFO methods are better able to optimize Link Divergence in the same number of epochs.

T-SNE\citep{vandermaaten08a} is a visualization technique that maps high-dimensional data to lower dimensions while preserving natural clusters. Using T-SNE, we observe that the GFO method produces node embeddings that do not naturally organize according to the value of the sensitive attribute in contrast to standard GCN autoencoder methods, which appear more visually separable under a T-SNE representation as shown in Figure \ref{fig:tsne}. Additionally, a T-SNE analysis of the learned $\bm{W}_f$ weights for GFO demonstrates that the method learns similar biases for nodes which share a sensitive attribute, as evidenced by the strong clustering of node-specific weight values observed in the left plot of Figure \ref{fig:tsne}.

Comparing the CFO$_{10}$ and CFO$_{100}$ methods, we observe that the CFO$_{100}$ method consistently achieves a lower $L_{D}$ loss and generally performs better than the CFO$_{10}$ method in regards to the fairness metrics. This comparison suggests that higher values of $c$ (the number of added nodes in CFO) are more capable of optimizing for fairness; this claim is further supported by the GFO performance, which can be considered a bound on the performance of CFO models due to the relation between the solution spaces for CFO and GFO fair weight matrices discussed in section \ref{sec:cfo}. To investigate this further, we run the CFO method with a wider range of values of $c$.

\subsection{Impact of the parameter $c$ in CFO}
\label{sec:cfo_c}
Results for a wider variety of CFO$_{c}$ models are shown in Figure \ref{fig:cfo_metrics} of Appendix \ref{app:cfo} for the Cora dataset. As $c$ increases, we see a slight increase in reconstruction loss; however, this does not correlate with an increase in AUROC, which steadily hovers around 0.73, indicating models with higher $c$ still perform well on the link prediction task. However, higher $c$ models generally exhibit stronger performance on the fairness-related metrics. As the number of introduced nodes $c$ increases, the Link Divergence loss decays exponentially, indicating CFO models with higher $c$ are more capable of optimizing for fairness. This is paralleled in the observed DP@40 metric, which similarly decreases as $c$ increases. After $c \approx 40$, Link Divergence and DP@40 stop decreasing significantly, indicating the number of introduced nodes has saturated and any further increases in $c$ will offer minimal returns for the Cora dataset.

\subsection{Efficiency of Approaches}

%start with time and memory complexity of GCN and then go through each method and how it changes it
% \yh{TODO: Review and edit for YH}

We first acknowledge that the time complexity of training a base $L$-layer GCN model is $O(L(|E|d + nd^{2}))$~\cite{wu2020comprehensive}, where $|E|$ is the number of edges, $d$ is the size of the embedding dimension, and $n$ is the number of nodes. In each layer, GCN performs $O(|E|)$ convolution operations with a sparse implementation then followed by a non-linear transformation, which takes $O(nd^2)$ time.

%\sfc{We first acknowledge that the time complexity of a GCN with sparse implementation is $O(|E|)$, where $|E|$ is the cardinality of the edge set $E$.}

The time complexity of our GFO approach remains $O(L(|E|d + nd^{2}))$. In each layer, GFO introduces an additive bias term into each node's representation after the convolution, which takes $O(nd)$ time. Since $n \ll |E|$, this computation does not increase the time complexity.

%\sfc{The GFO approach adds little computational cost to the neural network and does not significantly change the efficiency of the graph convolution neural network. In its simplest form, the GFO approach introduces an additive bias term into the graph convolution algorithm neural network architecture, which does not impact the big-$O$ algorithmic efficiency. The efficiency remains $O(|E|)$ for a sparse matrix implementation.}

% For the CFO approach, the time complexity is $O(L(max(|E|, cn)d + nd^{2}))$. CFO adds an additional $n \cdot c$ edges into the graph adjacency matrix, where $c$ is the number of additional nodes. With a sparse implementation of GCN, the additional $|E| + cn$ convolutions in each layer takes $O(ncd)$ time. When $c$ is not large, the total time complexity varies little compared to the base GCN.

For the CFO approach, the time complexity is $O(max(|E|, cn)d + (L-1)|E|d + Lnd^{2})$. CFO adds an additional $n \cdot c$ edges into the first layer of the graph adjacency matrix, where $c$ is the number of additional nodes. With a sparse implementation of GCN, the additional $cn$ convolutions in the first layer takes $O(ncd)$ time. When $c$ is not large, the total time complexity varies little compared to the base GCN.

%\sfc{When used properly, the CFO method should not significantly affect the efficiency of the graph convolution algorithm, though when taken to the extreme, the algorithm can impact runtime. Assuming matrices are stored in a sparse representation, the CFO approach adds an additional $n \times c$ edges into the computation, where $n$ is the number of existing nodes and $c$ is the number of additional nodes due to CFO. If the input graph $G$ has many edges and is highly connected, the number of edges $|E|$ outweighs the number of added edges $n \times c$ and is the main driving force of the efficiency of the algorithm. In contrast, if the number of added nodes $c$ is greater than the average number of edges each node in the graph has, then the CFO implementation is the limiting factor of the algorithmic efficiency. Thus, using the CFO method, the big-$O$ runtime is $O(\max(|E|, n \times c))$.}

Similar to the GFO approach, the FEW method does not affect the overall efficiency of the graph convolution algorithm. The FEW method introduces a weight for each edge in the input graph, which scale the adjacency matrix before the graph convolution step. This does not affect the big-$O$ runtime, which will remain $O(L(|E|d + nd^{2}))$ for a sparse matrix implementation.

\section{Conclusion}
\label{sec:conclusion}

Standard graph representation learning methods built for the task of link recommendation can learn node representations that are unfair and biased towards historically disadvantaged and underserved communities, leading to unfair treatment in ranking \cite{karimi2018homophily}, social perception \cite{lee2019homophily}, and job promotion \cite{tesch1995promotion, clifton2019mathematical}. In order to support equity and opportunity for said communities, we propose a set of three methods built to encourage diversity and equity in graph representation learning for link recommendation.

While many works have expanded the graph representation learning domain, few have considered fair graph representations and the graph modifications necessary to construct such embeddings. Existing works such as InFoRM \cite{kang2020inform}, FairWalk \cite{ijcai2019-456}, and FairAdj \citep{li2021dyadic} focus on introducing fairness by adjusting the input adjacency matrix; in contrast, our work emulates the effects of three unique graph modification methods, including methods that introduce new nodes into the graph, along with a novel loss function to address demographic parity in the graph embedding domain.

Notably, all three of our fairness methods demonstrate significant ability to improve demographic parity on the link recommendation task with negligible loss (and in some cases small gains) in link prediction performance. We show that for a set of four datasets, each formulation is able to increase graph fairness under the definition of demographic parity through effects similar to the introduction of new nodes or edge weights. Additionally, our methods are separable from the base GCN architecture used in the embedding learning, allowing users to extract information about the emulated graph modifications from the embedding model. In the future, we hope to demonstrate the flexibility of our formulations on a wider variety of GCN architectures, such as VGAE \cite{kipf2016variational}.

We additionally plan to investigate the types of modifications our methods learn and emulate, so that we may better understand how bias is represented and remedied in these datasets. We aspire to explore ways we might impose constraints on our fair learning methods to better simulate the introduction of realistic nodes and edges; while our current methods significantly improve demographic parity, introduced artificial nodes are free to take on whatever features will do the job, regardless of if those features could exist. Finally, we intend to construct new losses for optimization in order to utilize our emulated graph modification techniques to improve fairness on other embedding tasks, such as node classification, or for other forms of fairness, such as equalized odds or individual fairness.

\noindent {\bf Acknowledgements:}
This material is partially supported by the National Science Foundation (NSF) under grants OAC-2018627, CCF-2028944, and CNS-2112471 and the Air Force Office of Scientific Research  (AFOSR) under grant FA8650-19-2-2204. Any opinions, findings, and conclusions in this material are those of the author(s) and may not reflect the views of the respective funding agencies.

\bibliographystyle{ACM-Reference-Format}
\bibliography{mybib}

\newpage
\appendix

\newpage
\section{Augmented Loss Model}
\label{app:aug}

We present an additional experiment to compare the GFO, CFO, and FEW models to a standard GAE autoencoder trained using an augmented loss function of the form $L(\bm{\Phi}) = L_{R}(\bm{\Phi}) + \lambda L_{D}(\bm{\Phi})$. This modified loss function encourages the GAE model to learn both the utility and fairness tasks without introducing additional weights or terms into the GNN. Because the combined loss is highly sensitive to $\lambda$ for balancing utility and fairness, we test the AUG models with values of $\lambda = 10^i$ for $i=0,1,...,5$. This serves as an additional baseline for our models.

Tables \ref{tab:resultsaug1} and \ref{tab:resultsaug2} present the results of the augmented loss models (AUG$_{\lambda}$) on the utility and fairness metrics, respectively. In contrast to the GFO, CFO, and FEW models, the AUG model with $\lambda=1$ is not able to offer significantly fairer embeddings over the base GAE model, and offers only slight improvements in the fairness metrics while maintaining a similar degree of utility as the base model. However, as $\lambda$ increases, there is a noticeable shift in the balance of fairness and utility in the AUG models. Models with higher values of $\lambda$ tend to have stronger performance on fairness metrics.

Comparing the AUG to the GFO, CFO, and FEW models, we find that the AUG models perform similarly to GFO, CFO, and FEW on the link recommendation task, supporting earlier findings in Section \ref{sec:results} that the fairness optimizations learned by these models do not significantly impact embedding performance. In regards to fairness related performance, at least one of the GFO, CFO, and FEW models performs better or comparable to the AUG models on the Citeseer, Facebook, and Pubmed datasets. However, we do observe stronger performance for the AUG$_10000$ and AUG$_100000$ models on the Cora dataset for the DP\@10 and DP\@20 metrics. For DP\@40, the GFO and CFO$_{100}$ models are once again the strongest performers.

Overall, these results show that restricting the optimization of the Link Divergence loss to a specific set of weights in the GFO and CFO methods does not significantly impede the ability of these models to learn fairer embeddings. Additionally, these experimental results offer a fourth method of improving fairness with respect to demographic parity in graph embeddings via the augmented loss function $L(\bm{\Phi}) = L_{R}(\bm{\Phi}) + \lambda L_{D}(\bm{\Phi})$, which perform comparably or slightly better than the GFO, CFO, and FEW models at the cost of reduced interpretability.

\newpage
\begin{table}[!h]
    \centering
    \small
    \begin{tabular}{l l l l l l}
        \toprule
        Dataset & Model & $L_{R}$ $\downarrow$ & $L_{D}$ $\downarrow$ & AUROC $\uparrow$ & F1 $\uparrow$ \\
        \toprule
\multirow{11}{*}{Citeseer} & Base       & \textbf{1.31}       $\pm$ 0.0399     & 0.000741   $\pm$ 0.00246    & \textbf{0.74}       $\pm$ 0.0295     & \textbf{0.616}      $\pm$ 0.0261     \\
        \cmidrule{2-6}
& GFO        & 1.34       $\pm$ 0.0145     & 3.21e-08   $\pm$ 4.85e-08   & 0.714      $\pm$ 0.0263     & 0.599      $\pm$ 0.0261     \\
& CFO$_{10}$ & 1.34       $\pm$ 0.0125     & 9.88e-07   $\pm$ 9.14e-07   & 0.714      $\pm$ 0.0339     & 0.599      $\pm$ 0.0354     \\
& CFO$_{100}$ & 1.34       $\pm$ 0.014      & 6.7e-08    $\pm$ 6.65e-08   & 0.706      $\pm$ 0.0544     & 0.596      $\pm$ 0.0339     \\
& FEW        & 1.33       $\pm$ 0.0242     & 0.000124   $\pm$ 0.000161   & 0.718      $\pm$ 0.0579     & 0.608      $\pm$ 0.0345     \\
        \cmidrule{2-6}
& AUG$_{1}$  & 1.33       $\pm$ 0.0237     & 0.000121   $\pm$ 0.000133   & 0.703      $\pm$ 0.0687     & 0.602      $\pm$ 0.0377     \\
& AUG$_{10}$ & 1.33       $\pm$ 0.0208     & 0.000105   $\pm$ 0.000106   & 0.714      $\pm$ 0.0604     & 0.605      $\pm$ 0.0339     \\
& AUG$_{100}$ & 1.34       $\pm$ 0.0187     & 2.29e-05   $\pm$ 1.6e-05    & 0.71       $\pm$ 0.053      & 0.597      $\pm$ 0.0278     \\
& AUG$_{1000}$ & 1.34       $\pm$ 0.0128     & 2.54e-06   $\pm$ 1.34e-06   & 0.707      $\pm$ 0.0518     & 0.597      $\pm$ 0.0252     \\
& AUG$_{10000}$ & 1.35       $\pm$ 0.014      & 1.78e-07   $\pm$ 1.39e-07   & 0.704      $\pm$ 0.057      & 0.599      $\pm$ 0.0345     \\
& AUG$_{100000}$ & 1.35       $\pm$ 0.0195     & \textbf{2.29e-08}   $\pm$ 2.17e-08   & 0.665      $\pm$ 0.0932     & 0.584      $\pm$ 0.036      \\
        \toprule
\multirow{11}{*}{Cora} & Base       & \textbf{1.29}       $\pm$ 0.0188     & 0.000117   $\pm$ 0.000159   & 0.73       $\pm$ 0.0205     & 0.608      $\pm$ 0.0199     \\
        \cmidrule{2-6}
& GFO        & 1.31       $\pm$ 0.00939    & \textbf{1.04e-07}   $\pm$ 6.5e-08    & \textbf{0.731}      $\pm$ 0.0182     & \textbf{0.613}      $\pm$ 0.0183     \\
& CFO$_{10}$ & 1.3        $\pm$ 0.0155     & 7.71e-06   $\pm$ 5.44e-06   & \textbf{0.731}      $\pm$ 0.0204     & \textbf{0.613}      $\pm$ 0.0188     \\
& CFO$_{100}$ & 1.3        $\pm$ 0.0116     & 4.88e-07   $\pm$ 7.2e-07    & 0.729      $\pm$ 0.0207     & 0.609      $\pm$ 0.022      \\
& FEW        & \textbf{1.29}       $\pm$ 0.0188     & 0.000101   $\pm$ 0.000112   & 0.726      $\pm$ 0.0226     & 0.607      $\pm$ 0.0242     \\
        \cmidrule{2-6}
& AUG$_{1}$  & \textbf{1.29}       $\pm$ 0.0165     & 9.75e-05   $\pm$ 7.34e-05   & 0.73       $\pm$ 0.0186     & 0.613      $\pm$ 0.0214     \\
& AUG$_{10}$ & \textbf{1.29}       $\pm$ 0.0227     & 0.000109   $\pm$ 9.66e-05   & 0.73       $\pm$ 0.0192     & 0.606      $\pm$ 0.0183     \\
& AUG$_{100}$ & 1.3        $\pm$ 0.0136     & 3.68e-05   $\pm$ 1.95e-05   & 0.727      $\pm$ 0.0196     & 0.611      $\pm$ 0.0203     \\
& AUG$_{1000}$ & 1.3        $\pm$ 0.0145     & 9.53e-06   $\pm$ 5.09e-06   & 0.728      $\pm$ 0.0239     & 0.606      $\pm$ 0.0236     \\
& AUG$_{10000}$ & 1.32       $\pm$ 0.00953    & 1.52e-06   $\pm$ 9.64e-07   & 0.729      $\pm$ 0.0195     & 0.61       $\pm$ 0.0206     \\
& AUG$_{100000}$ & 1.32       $\pm$ 0.0106     & 2.05e-07   $\pm$ 1.57e-07   & 0.727      $\pm$ 0.0216     & 0.603      $\pm$ 0.0165     \\
        \toprule
\multirow{11}{*}{Facebook} & Base       & 1.25       $\pm$ 0.0295     & 1.43e-05   $\pm$ 1.65e-05   & 0.786      $\pm$ 0.0117     & 0.72       $\pm$ 0.0116     \\
        \cmidrule{2-6}
& GFO        & 1.27       $\pm$ 0.0189     & 6.89e-08   $\pm$ 5.99e-08   & 0.788      $\pm$ 0.0094     & 0.72       $\pm$ 0.00813    \\
& CFO$_{10}$ & 1.26       $\pm$ 0.0185     & 4.45e-07   $\pm$ 3.96e-07   & 0.786      $\pm$ 0.00998    & 0.72       $\pm$ 0.00944    \\
& CFO$_{100}$ & 1.26       $\pm$ 0.0177     & \textbf{3.01e-08}   $\pm$ 1.66e-08   & 0.787      $\pm$ 0.00966    & 0.721      $\pm$ 0.00704    \\
& FEW        & \textbf{1.24}       $\pm$ 0.0409     & 1.65e-05   $\pm$ 2.34e-05   & 0.787      $\pm$ 0.00998    & 0.721      $\pm$ 0.00921    \\
        \cmidrule{2-6}
& AUG$_{1}$  & 1.25       $\pm$ 0.0308     & 1.38e-05   $\pm$ 1.18e-05   & 0.788      $\pm$ 0.00639    & 0.721      $\pm$ 0.00508    \\
& AUG$_{10}$ & \textbf{1.24}       $\pm$ 0.0325     & 1.65e-05   $\pm$ 1.33e-05   & 0.787      $\pm$ 0.00911    & 0.721      $\pm$ 0.00746    \\
& AUG$_{100}$ & \textbf{1.24}       $\pm$ 0.0455     & 2.19e-05   $\pm$ 4.08e-05   & 0.787      $\pm$ 0.00953    & 0.721      $\pm$ 0.00897    \\
& AUG$_{1000}$ & 1.25       $\pm$ 0.0256     & 8.77e-06   $\pm$ 8.68e-06   & 0.788      $\pm$ 0.00907    & 0.721      $\pm$ 0.00942    \\
& AUG$_{10000}$ & 1.27       $\pm$ 0.0122     & 8.94e-07   $\pm$ 3.53e-07   & \textbf{0.791}      $\pm$ 0.00893    & \textbf{0.724}      $\pm$ 0.00678    \\
& AUG$_{100000}$ & 1.28       $\pm$ 0.0133     & 1.15e-07   $\pm$ 9.5e-08    & 0.788      $\pm$ 0.00686    & 0.723      $\pm$ 0.00444    \\
        \toprule
\multirow{11}{*}{Pubmed} & Base       & 1.31       $\pm$ 0.00417    & 2.95e-06   $\pm$ 2.72e-06   & \textbf{0.847}      $\pm$ 0.00431    & \textbf{0.724}      $\pm$ 0.00499    \\
        \cmidrule{2-6}
& GFO        & 1.31       $\pm$ 0.0179     & 2.22e-09   $\pm$ 3.17e-09   & 0.829      $\pm$ 0.0717     & 0.716      $\pm$ 0.0314     \\
& CFO$_{10}$ & 1.31       $\pm$ 0.0181     & 4.56e-09   $\pm$ 3.44e-09   & 0.83       $\pm$ 0.0758     & 0.718      $\pm$ 0.0323     \\
& CFO$_{100}$ & 1.32       $\pm$ 0.0242     & \textbf{1.84e-09}   $\pm$ 1.5e-09    & 0.812      $\pm$ 0.104      & 0.711      $\pm$ 0.0444     \\
& FEW        & 1.31       $\pm$ 0.0176     & 5.41e-08   $\pm$ 7.84e-08   & 0.834      $\pm$ 0.053      & 0.716      $\pm$ 0.0318     \\
        \cmidrule{2-6}
& AUG$_{1}$  & 1.32       $\pm$ 0.0241     & 2.68e-06   $\pm$ 2.83e-06   & 0.81       $\pm$ 0.104      & 0.708      $\pm$ 0.0428     \\
& AUG$_{10}$ & 1.31       $\pm$ 0.00723    & 3.54e-06   $\pm$ 3.3e-06    & 0.846      $\pm$ 0.00606    & \textbf{0.724}      $\pm$ 0.00671    \\
& AUG$_{100}$ & 1.32       $\pm$ 0.0238     & 2.36e-06   $\pm$ 4.04e-06   & 0.81       $\pm$ 0.104      & 0.709      $\pm$ 0.0446     \\
& AUG$_{1000}$ & 1.31       $\pm$ 0.0244     & 2.93e-07   $\pm$ 2.36e-07   & 0.813      $\pm$ 0.104      & 0.71       $\pm$ 0.0434     \\
& AUG$_{10000}$ & \textbf{1.3}        $\pm$ 0.00639    & 4.24e-08   $\pm$ 4.45e-08   & \textbf{0.847}      $\pm$ 0.00438    & \textbf{0.724}      $\pm$ 0.00496    \\
& AUG$_{100000}$ & 1.32       $\pm$ 0.0236     & 1.12e-08   $\pm$ 9.8e-09    & 0.816      $\pm$ 0.0923     & 0.71       $\pm$ 0.0418     \\
        \bottomrule
    \end{tabular}
    \captionsetup{justification=centering}
    \caption{Utility in Link Prediction: losses and link prediction metrics for the augmented loss model on all datasets. Prior results for the GFO, CFO, and FEW methods are repeated for ease of comparison. The highest performing model for each dataset and metric is bolded.}
    \label{tab:resultsaug1}
\end{table}

\newpage
\begin{table}[!h]
    \centering
    \small
    \begin{tabular}{l l l l l}
        \toprule
        Dataset & Model & DP@10 $\downarrow$ & DP@20 $\downarrow$ & DP@40 $\downarrow$ \\
        \toprule
\multirow{11}{*}{Citeseer} & Base       & 4.88       $\pm$ 1.79       & 3.01       $\pm$ 1.69       & 1.85       $\pm$ 1.4        \\
        \cmidrule{2-5}
& GFO        & 2.3        $\pm$ 1.23       & \textbf{0.703}      $\pm$ 0.497      & \textbf{0.17}       $\pm$ 0.197      \\
& CFO$_{10}$ & \textbf{1.94}       $\pm$ 0.791      & 0.879      $\pm$ 0.527      & 0.234      $\pm$ 0.285      \\
& CFO$_{100}$ & 2.35       $\pm$ 1.17       & 0.71       $\pm$ 0.692      & 0.271      $\pm$ 0.516      \\
& FEW        & 4.17       $\pm$ 1.35       & 2.2        $\pm$ 1.28       & 1.17       $\pm$ 0.937      \\
        \cmidrule{2-5}
& AUG$_{1}$  & 4.28       $\pm$ 1.35       & 2.26       $\pm$ 1.27       & 1.38       $\pm$ 1.12       \\
& AUG$_{10}$ & 4.76       $\pm$ 1.87       & 2.76       $\pm$ 1.86       & 1.57       $\pm$ 1.34       \\
& AUG$_{100}$ & 3.15       $\pm$ 1.33       & 1.65       $\pm$ 0.902      & 0.593      $\pm$ 0.433      \\
& AUG$_{1000}$ & 2.25       $\pm$ 0.892      & 0.98       $\pm$ 0.8        & 0.398      $\pm$ 0.585      \\
& AUG$_{10000}$ & 2.09       $\pm$ 0.962      & 0.861      $\pm$ 0.667      & 0.395      $\pm$ 0.504      \\
& AUG$_{100000}$ & 2.11       $\pm$ 1.39       & 0.933      $\pm$ 0.993      & 0.574      $\pm$ 0.802      \\
        \toprule
\multirow{11}{*}{Cora} & Base       & 3.79       $\pm$ 1.06       & 1.89       $\pm$ 0.934      & 0.914      $\pm$ 0.652      \\
        \cmidrule{2-5}
& GFO        & 3.47       $\pm$ 0.826      & 1.05       $\pm$ 0.348      & \textbf{0.132}      $\pm$ 0.0486     \\
& CFO$_{10}$ & 3.38       $\pm$ 0.84       & 1.23       $\pm$ 0.536      & 0.239      $\pm$ 0.123      \\
& CFO$_{100}$ & 3.32       $\pm$ 0.808      & 0.96       $\pm$ 0.424      & 0.147      $\pm$ 0.0716     \\
& FEW        & 3.81       $\pm$ 0.874      & 1.87       $\pm$ 0.618      & 0.719      $\pm$ 0.42       \\
        \cmidrule{2-5}
& AUG$_{1}$  & 3.9        $\pm$ 0.925      & 1.82       $\pm$ 0.8        & 0.798      $\pm$ 0.552      \\
& AUG$_{10}$ & 4.12       $\pm$ 0.966      & 2.05       $\pm$ 0.84       & 0.918      $\pm$ 0.569      \\
& AUG$_{100}$ & 3.64       $\pm$ 0.654      & 1.37       $\pm$ 0.531      & 0.454      $\pm$ 0.284      \\
& AUG$_{1000}$ & 3.65       $\pm$ 0.861      & 1.28       $\pm$ 0.45       & 0.237      $\pm$ 0.132      \\
& AUG$_{10000}$ & 3.24       $\pm$ 1.1        & 1.06       $\pm$ 0.695      & 0.178      $\pm$ 0.194      \\
& AUG$_{100000}$ & \textbf{2.76}       $\pm$ 1.14       & \textbf{0.862}      $\pm$ 0.539      & 0.206      $\pm$ 0.277      \\
        \toprule
\multirow{11}{*}{Facebook} & Base       & 0.136      $\pm$ 0.161      & 0.0418     $\pm$ 0.0204     & 0.0226     $\pm$ 0.0102     \\
        \cmidrule{2-5}
& GFO        & \textbf{0.0661}     $\pm$ 0.0484     & \textbf{0.0191}     $\pm$ 0.0114     & 0.0076     $\pm$ 0.00446    \\
& CFO$_{10}$ & 0.17       $\pm$ 0.329      & 0.0226     $\pm$ 0.00984    & \textbf{0.0072}     $\pm$ 0.00366    \\
& CFO$_{100}$ & 0.101      $\pm$ 0.0643     & 0.0246     $\pm$ 0.0132     & 0.00783    $\pm$ 0.00441    \\
& FEW        & 0.217      $\pm$ 0.391      & 0.0413     $\pm$ 0.0252     & 0.0171     $\pm$ 0.00914    \\
        \cmidrule{2-5}
& AUG$_{1}$  & 0.103      $\pm$ 0.0815     & 0.0416     $\pm$ 0.0245     & 0.0251     $\pm$ 0.0116     \\
& AUG$_{10}$ & 0.109      $\pm$ 0.0552     & 0.042      $\pm$ 0.0253     & 0.0249     $\pm$ 0.013      \\
& AUG$_{100}$ & 0.0987     $\pm$ 0.0549     & 0.0365     $\pm$ 0.022      & 0.0245     $\pm$ 0.0131     \\
& AUG$_{1000}$ & 0.108      $\pm$ 0.0697     & 0.036      $\pm$ 0.0208     & 0.0195     $\pm$ 0.00992    \\
& AUG$_{10000}$ & 0.148      $\pm$ 0.345      & 0.026      $\pm$ 0.0168     & 0.00878    $\pm$ 0.00496    \\
& AUG$_{100000}$ & 0.179      $\pm$ 0.229      & 0.0324     $\pm$ 0.0279     & 0.00978    $\pm$ 0.00621    \\
        \toprule
\multirow{11}{*}{Pubmed} & Base       & 2.43       $\pm$ 1.36       & 0.64       $\pm$ 1.09       & 0.185      $\pm$ 0.0859     \\
        \cmidrule{2-5}
& GFO        & \textbf{1.52}       $\pm$ 1.56       & 0.327      $\pm$ 0.72       & 0.196      $\pm$ 0.0851     \\
& CFO$_{10}$ & 2.61       $\pm$ 1.27       & \textbf{0.174}      $\pm$ 0.0407     & 0.203      $\pm$ 0.0694     \\
& CFO$_{100}$ & 2.06       $\pm$ 1.51       & 0.182      $\pm$ 0.0322     & 0.168      $\pm$ 0.0906     \\
& FEW        & 2.31       $\pm$ 1.38       & 0.328      $\pm$ 0.697      & 0.196      $\pm$ 0.0686     \\
        \cmidrule{2-5}
& AUG$_{1}$  & 1.66       $\pm$ 1.51       & 0.507      $\pm$ 0.865      & \textbf{0.165}      $\pm$ 0.0907     \\
& AUG$_{10}$ & 2.78       $\pm$ 1.13       & 0.735      $\pm$ 1.19       & 0.225      $\pm$ 0.0776     \\
& AUG$_{100}$ & 2.48       $\pm$ 1.36       & 0.331      $\pm$ 0.69       & 0.19       $\pm$ 0.0914     \\
& AUG$_{1000}$ & 2.22       $\pm$ 1.44       & 0.323      $\pm$ 0.498      & 0.217      $\pm$ 0.0705     \\
& AUG$_{10000}$ & 2.5        $\pm$ 1.35       & 0.485      $\pm$ 0.924      & 0.207      $\pm$ 0.0744     \\
& AUG$_{100000}$ & 2.27       $\pm$ 1.37       & 0.672      $\pm$ 1.04       & 0.195      $\pm$ 0.0712     \\
        \bottomrule
    \end{tabular}
    \captionsetup{justification=centering}
    \caption{Fairness in Link Recommendations: DP@$k$ metrics for the augmented loss model on all datasets. Prior results for the GFO, CFO, and FEW methods are repeated for ease of comparison. The highest performing model for each dataset and metric is bolded.}
    \label{tab:resultsaug2}
\end{table}

\newpage
\section{Additional Results for CFO}
\label{app:cfo}

We present an additional experiment to test the impact of the number of introduced nodes $c$ in the CFO method, as discussed in section \ref{sec:cfo_c}. Experiments were run over 300 epochs using 5-fold cross validation. As $c$ increases, the achieved link divergence and DP@40 metrics decay exponentially while the reconstruction loss trends upward slightly. There is no consistent change in the average AUROC score with increased $c$.

\begin{figure*}[!h]
    \centering
    \includegraphics[width=\textwidth]{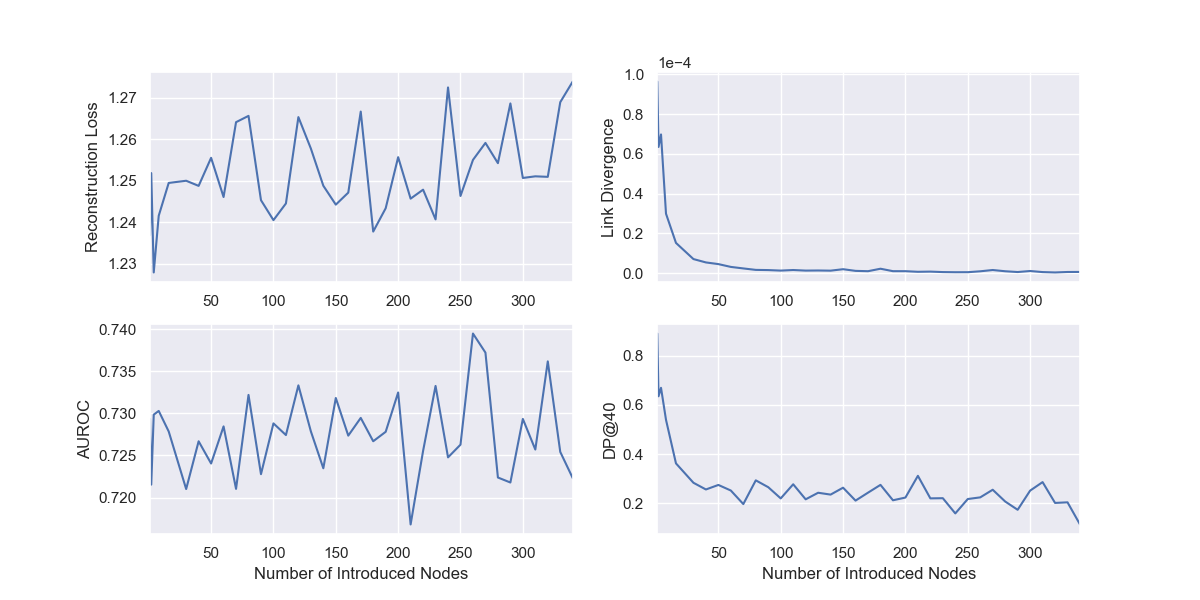}
    \caption{Top left: Reconstruction loss of the CFO$_{c}$ model for varying values of $c$. Top right: Link divergence of the CFO$_{c}$ model for varying values of $c$. Bottom left: AUROC metric of the CFO$_{c}$ model for varying values of $c$. Bottom right: DP@40 metric of the CFO$_{c}$ model for varying values of $c$. All models were trained for 300 epochs with 5-fold cross validation on the Cora dataset. Values of $c$ taken into consideration were $1, 2, 4, 8, 16, 30, 40,...,340$.}
    \label{fig:cfo_metrics}
\end{figure*}

\end{document}